\documentclass[runningheads]{llncs}
\usepackage[T1]{fontenc}
\usepackage[utf8]{inputenc}
\usepackage{graphicx}
\usepackage{amsfonts,amssymb}
\usepackage{color}
\usepackage{algorithm}
\usepackage[noend]{algpseudocode}
\usepackage{caption}
\usepackage{subcaption}
\usepackage{array}
\usepackage{amsmath,amsfonts,amssymb}
\usepackage{graphicx}
\usepackage{setspace}
\usepackage{tocloft}
\usepackage{threeparttable}
\usepackage{lineno}
\usepackage{booktabs}
\usepackage{upgreek}
\usepackage{multirow}
\usepackage{lineno}

\usepackage[breaklinks=true]{hyperref}
\usepackage{breakurl}
\usepackage{cite}
\DeclareUnicodeCharacter{02BC}{'}
\hypersetup{
    colorlinks=true,
    linkcolor=blue,
    citecolor=blue,
    filecolor=blue,
    urlcolor=blue,
    pdfborder={0 0 0} %
}

\begin{document}
\title{Benchmarking DINOv3 for Multi-Task Stroke Analysis on Non-Contrast CT}
%
%

\author{Donghao Zhang\inst{1}  \and
Yimin Chen\inst{1} \and
Kauê TN Duarte\inst{3} \and
Taha Aslan\inst{3} \and
Mohamed AlShamrani\inst{3} \and
Brij Karmur\inst{3} \and
Yan Wan\inst{4} \and
Shengcai Chen\inst{4} \and
Bo Hu\inst{4} \and
Bijoy K Menon\inst{3} \and
Wu Qiu\inst{1,2,*}
 }

\authorrunning{Donghao Zhang et al.}
%

\institute{Department of Biomedical Engineering, School of Life Science and Technology, Huazhong University of Science and Technology, Wuhan 430074, People's Republic of China\\
\email{Qiu.wu.ch@gmail.com} \and
Advanced Biomedical Imaging Facility, HUST, Wuhan, 430074, Peopleʼs Republic of China\\ \and
Departments of Clinical Neuroscience and Radiology, Hotchkiss Brain Institute, Cummings School of Medicine, University of Calgary,Calgary,AB, Canada\\ \and
Union Hospital, Tongji Medical College, Huazhong University of Science and Technology, Wuhan, China\\ 
}
\maketitle             

\negthinspace\negthinspace\negthinspace\negthinspace
\begin{abstract}
\negthinspace

Non-contrast computed tomography (NCCT) is essential for rapid stroke diagnosis but is limited by low image contrast and signal to noise ratio. We address this challenge by leveraging DINOv3, a state-of-the-art self-supervised vision transformer, to generate powerful feature representations for a comprehensive set of stroke analysis tasks. Our evaluation encompasses infarct and hemorrhage segmentation, anomaly classification (normal vs.\ stroke and normal vs.\ infarct vs.\ hemorrhage), hemorrhage subtype classification (EDH, SDH, SAH, IPH, IVH), and dichotomized ASPECTS classification (<=6 vs. >6) on multiple public and private datasets. This study establishes strong benchmarks for these tasks and demonstrates the potential of advanced self-supervised models to improve automated stroke diagnosis from NCCT, providing a clear analysis of both the advantages and current constraints of the approach. The code is available at \href{https://github.com/Zzz0251/DINOv3-stroke}{https://github.com/Zzz0251/DINOv3-stroke}.

\negthinspace\negthinspace
\end{abstract}

\section{Introduction}
\negthinspace

Stroke remains one of the leading causes of mortality and long-term disability worldwide. Non-contrast computed tomography (NCCT) is the primary imaging modality in the acute setting due to its speed, accessibility, and non-invasive nature\cite{strokeReview1}. However, NCCT suffers from inherently low contrast and subtle early stroke changes, which makes both clinical interpretation and automated analysis challenging. Existing machine learning approaches have shown promising results in stroke-related classification and lesion segmentation, but their performance on NCCT remains limited, including  generalization\cite{WANG2025108488,UMAMAHESWARAN2024122559,NEETHI2022103720}. This is largely due to the difficulty of accurately representing complex brain structures and subtle stroke manifestations.

Recently, large-scale self-supervised vision models have demonstrated remarkable semantic understanding across diverse domains\cite{visionR10834497,visionR3,VisionR10916803}. Among them, DINOv3\cite{simeoni2025dinov3} represents a state-of-the-art vision transformer, pretrained on more than 1.6 billion images, and capable of extracting high-fidelity, fine-grained dense features. These properties suggest that DINOv3 may help overcome key limitations of NCCT-based stroke analysis by providing more robust representations of brain anatomy and subtle abnormalities.

In this study, we systematically evaluate DINOv3 for NCCT-based stroke analysis across multiple tasks, including infarct and hemorrhage segmentation, anomaly classification, hemorrhage subtype classification, and ASPECTS score classification. Our results aim to provide a basic systematic assessment of the capabilities and limitations of large-scale self-supervised vision models for stroke diagnosis on NCCT, offering a reference point for future research in this domain.

\section{Related Work}
\negthinspace

Early studies on NCCT-based stroke analysis mainly relied on handcrafted features and radiomics approaches\cite{11078788,ZHANG2023110959}. While these methods offered some utility for detecting hemorrhage or quantifying ischemic changes, their effectiveness was limited by the low contrast of NCCT and the subtle presentation of early stroke signs. 

With the rise of deep learning, convolutional neural networks (CNNs) such as U-Net and its variants became the dominant paradigm for medical image segmentation, including infarct and hemorrhage detection. Vision transformers have also been explored for medical imaging, offering improved ability to capture long-range dependencies\cite{ZHANG2023110959}. However, these methods are usually fully supervised and require large annotated datasets, which are difficult to obtain in practice. Moreover, their performance has been stronger on advanced modalities such as MRI, CTA, or CTP, but less reliable on NCCT.

To reduce reliance on annotated data, self-supervised learning has emerged as a promising direction in medical imaging. Contrastive frameworks (e.g., SimCLR\cite{pmlr-v119-chen20j}, CLIP\cite{radford2021learning}), masked image modeling (e.g., MAE\cite{he2021masked}) have shown encouraging results by learning general-purpose representations from unlabeled data. Nonetheless, prior work remains limited in scale and scope, and systematic evaluations of large self-supervised models for NCCT stroke analysis are still lacking. Against this background, our work provides the first comprehensive benchmark of DINOv3, a state-of-the-art self-supervised vision transformer, for both stroke classification and lesion segmentation tasks on NCCT.

\section{Methods}

\subsection{Overview}
\negthinspace
This study investigates the transferability of DINOv3, a state-of-the-art self-supervised vision transformer, to NCCT-based stroke-related tasks. DINOv3 represents a significant milestone in large-scale self-supervised learning: it was trained on over 1689M images (LVD-1689M) without the need for manual annotations, and demonstrates exceptional generalization across diverse downstream tasks. Compared with its predecessor, DINOv3  introduces a masked image modeling objective inspired by iBOT\cite{zhou2022ibot}, KoLeo regularization on the [CLS] token, and Gram anchoring to mitigate dense feature map degradation during long training. Additionally, DINOv3 adopts axial rotary position embeddings (RoPE)\cite{su2021roformer} with position regularization to mitigate positional artifacts. These improvements allow DINOv3 to produce high-quality dense features that remain effective under frozen-backbone settings, enabling efficient adaptation with lightweight task-specific heads. 

We adopt the distilled ViT-B version of DINOv3 as the backbone, which is kept frozen for all experiments. Only lightweight, task-specific heads are trained for classification or segmentation.

\subsection{Classification Tasks}
\negthinspace
For classification, we adopt a transfer learning strategy using features from the frozen DINOv3-ViT-B encoder, on top of which a task-specific head is trained. In the 2D setting, features from individual NCCT slices are passed through a linear head for binary tasks, while a deeper head (512$\rightarrow$256$\rightarrow n_{\text{class}}$) is employed for multi-class problems such as three- or five-class classification. For 3D NCCT scans, each volume is represented by 16 sampled slices, and the slice-wise features are aggregated into a single embedding via mean pooling fusion. This aggregated representation is then processed by the same multi-layer head (512$\rightarrow$256$\rightarrow n_{\text{class}}$) for final prediction.

\subsection{Segmentation Tasks}
\negthinspace
To evaluate the transferability of DINOv3 features for lesion segmentation, we experiment with both 2D and 3D decoder designs. In the slice-level case, three decoder variants are compared. One option is an upsampling decoder in the style of U-Net\cite{unet}, which employs convolutional and upsampling layers without skip connections. Another strategy leverages multi-layer feature fusion, where features from multiple transformer layers (e.g., [2,5,8,11], [1,3,5,7,9,11], or [1--11]) are concatenated and passed through an MLP-based head to capture multi-scale context. A third approach follows a DPT-style design\cite{yang2025segdino}, projecting and refining features from several layers before hierarchical fusion to integrate global context with local detail recovery. Extending to the 3D case, slice embeddings are first projected from 768 to 128 dimensions and stacked along the slice axis to form a volumetric tensor, which is then decoded by a lightweight 3D convolutional module with two convolutional blocks and upsampling layers. This design reconstructs volumetric lesion masks while capturing inter-slice dependencies without relying on encoder skip connections.

\subsection{Evaluation Metrics}
\negthinspace
For classification tasks, we report AUC, accuracy (ACC), sensitivity (SEN), and specificity (SPE). For segmentation tasks, we report Dice similarity coefficient (Dice), intersection over union (IoU) and HD95. These metrics are commonly used in classification and segmentation tasks. We employ these metrics to facilitate comparisons with previous and future related work.

\section{Clinical Data for NCCT-Based Stroke Diagnosis and Prognosis}
\negthinspace
In this study, we exclusively use non-contrast computed tomography (NCCT) images as the input modality for all tasks. NCCT is widely used in stroke diagnosis due to its speed, accessibility, and non-invasive nature\cite{Eswaradass02082016}. To ensure generalizability, we incorporate datasets from diverse imaging devices and patient populations, including publicly available and private datasets. The datasets cover clinically relevant classification and segmentation tasks, reflecting the workflow of stroke diagnosis. Below, we describe the datasets grouped by task type.

\subsection{Stroke Classification}
\negthinspace
In the clinical workflow, initial classification of NCCT scans determines whether a patient has a stroke and, if so, whether it is hemorrhagic or ischemic. This classification guides treatment decisions\cite{strokeReview1}. We evaluate this step using multiple datasets:

\noindent\textbf{Binary Classification: HeadCT Dataset\cite{Anjum_HeadCT}}  
This publicly available dataset contains 200 NCCT slices, with 100 labeled as normal and 100 as hemorrhagic. It is used to identify hemorrhagic strokes versus normal scans. This dataset is part of \cite{Anjum_HeadCT}, and we use the publicly available portion from Kaggle\footnote{\url{https://www.kaggle.com/datasets/felipekitamura/head-ct-hemorrhage}}.

\noindent\textbf{Multiclass Classification: Private Dataset}  
A dataset of 13,437 slices classified into three categories: normal (4,244), infarct (6,508, partly from AISD\cite{liang2021SymmetryEnhancedAN}), and hemorrhage (2,685). Infarct cases display clearly visible hypodense regions, and hemorrhage cases include multiple subtypes. Images are normalized to reduce bias from acquisition differences.

\noindent\textbf{Hemorrhage Subtype Classification: BHSD Dataset\cite{wu2023bhsd}}  
192 NCCT scans with annotations for subtypes (EDH, SDH, SAH, IPH, IVH) and pixel-level segmentation, plus 1,980 unlabeled scans. Used for multi-class hemorrhage subtype classification. We use 192 NCCT scans with detailed annotations for hemorrhage types classification and 5-classes segmentation. These 192 scans have detailed segmentation labels for five types of hemorrhage subtypes. We can not only use it to perform segmentation tasks for each of the five subtypes, but also use this label to conduct classification tasks on which slices have which type of hemorrhage.

\noindent\textbf{ASPECTS Scoring (Baseline NCCT)}  
To assess early ischemic changes, we formulate ASPECTS scoring as a binary classification task: 0–6 (\emph{extensive early ischemic change}) vs 7–10 (\emph{limited early ischemic change}). This task uses 313 baseline NCCT scans (train: 27 extensive / 223 limited; test: 7 extensive / 56 limited). The dataset is used for reflecting clinically relevant assessment of early ischemia.

\begin{table}[ht]
\centering
\caption{NCCT datasets for classification tasks. P = Public, Pr = Private.}
\label{tab:cls_datasets}
\begin{tabular}{l l l l}
\hline
Dataset & Pr/P & Samples (Train / Test) & Classes \\
\hline
HeadCT & P & 200 slices (100 / 100) & Normal/Hemorrhage \\
Private 3-class & Pr & 13437 slices (10752 / 2685) & Normal/Infarct/Hemorrhage \\
BHSD Subtypes & P & 2368 slices (1894 / 474) & EDH/SDH/SAH/IPH/IVH \\
ASPECTS & Pr & 313 scans (250 / 63) & 0–6/7–10 \\
\hline
\end{tabular}
\end{table}

\begin{table}[ht]
\centering
\caption{NCCT datasets for segmentation tasks. P = Public, Pr = Private.}
\label{tab:seg_datasets}
\begin{tabular}{l l l l}
\hline
Dataset & Pr/P & Samples (Train / Test) & Labels \\
\hline
BHSD & P & 2368 slices (1894 / 474) & 5 type Hemorrhage masks \\
IHDS & P & 318 slices (254 / 64) & Hemorrhage masks \\
Private Hemorrhage & Pr & 2685 slices (1876 / 809) & Hemorrhage masks \\
AISD & P & 3786 slices (2945 / 841) & Infarct masks \\
ISLES’24 & P & 3982 slices (3185 / 797) & Infarct masks \\
Private Infarct & Pr & 7,338 slices (5737 / 1601) & Infarct masks \\
Private 3D Infarct & Pr & 249 cases (200 / 49) & 3D infarct masks \\
\hline
\end{tabular}
\end{table}

\subsection{Lesion Segmentation}
\negthinspace
Once a stroke is identified, delineating lesions is critical for prognosis and treatment planning. We perform segmentation for both hemorrhage and infarct lesions using the following datasets:
\subsubsection{Hemorrhage Segmentation}
\negthinspace

\noindent\textbf{BHSD Dataset\cite{wu2023bhsd}}  
In this dataset, there are 192 scans with detailed segmentation results of five types of bleeding, each labeled in different colors. We have isolated all the slices with labels and conducted separate five-category segmentation tasks.

\noindent\textbf{IHDS Dataset\cite{Hssayeni2020CTHemorrhage}\cite{data5010014}}  
This dataset contains 82 CT scans. Among them, 36 scans are of patients with cerebral hemorrhage, but no specific distinction is made among the five types. For these 36 data points, only a complete segmentation label for the hemorrhage lesion is provided. We extracted the slices with labels and obtained 318 slices. These were randomly divided into 254 for training and 64 for testing. Although the dataset is small, we can still observe the excellent performance of the model.

\noindent\textbf{Private Dataset}  
We ourselves organized a private dataset for performing blood segmentation. This dataset did not classify specific types of bleeding but did contain various types of bleeding, with only a label indicating the location of the lesion. This dataset had 2685 slices, and we randomly split them into 1876 for training and 809 for testing.

\subsubsection{Infarct Segmentation}
\negthinspace

\noindent\textbf{AISD Dataset\cite{liang2021SymmetryEnhancedAN}}  
AISD is an open dataset widely used in the field of ischemic stroke segmentation. This dataset contains 397 NCCT scans. The labels were extracted from the DWI scans made 24 hours after the CT. We processed the three-dimensional data and turned it into 3786 slices. Using all the slices with labels, we randomly divided them into 2945 for training and 841 for testing.

\noindent\textbf{ISLES’24 Dataset\cite{Riedel2024ISLES}\cite{Rosa2024ISLES}}  
ISLES'24 is a dataset from a MICCAI challenge in 2024. This dataset contains data from various imaging modalities, including NCCT, CTA, CTP, MRI, as well as some clinical text information, which is very comprehensive. We only used the currently available data that could be downloaded. Among them, there were 149 NCCT data. We also extracted the labeled slices from this dataset to form our dataset. Finally, we compiled 3982 slices, randomly dividing them into 3185 for training and 797 for testing.

\noindent\textbf{Private Dataset}  
At the same time, we also used our own private data to create a dataset for infarct segmentation based on slices. The segmentation results of this dataset are somewhat on the large side. Not only do they include infarcts, but they also encompass the more difficult-to-distinguish and potentially salvageable penumbra areas around them. Our dataset contains 7338 slices. The 5737 slices were randomly split for training, and 1601 were used for testing.

\noindent\textbf{Volumetric Infarct Segmentation}  
To better capture spatial extent, a private dataset of 249 cases with 3D infarct masks co-registered to baseline NCCT is used (train: 200 / test: 49). This task assesses infarct burden and spatial distribution, integrating volumetric context naturally with the segmentation task.

\section{Experiments and Results}
\negthinspace

\subsection{Experiments Implementation}
\negthinspace

All experiments are conducted using a official frozen DINOv3-ViT-B backbone distilled from the 7B model, with only task-specific heads trained to assess feature transferability. For classification, either a linear or a shallow MLP head is used; for segmentation, slice-based decoders are employed, including simple upsampling, multi-layer feature fusion, and a DPT-style (Dense Prediction Transformer) FPN (Feature Pyramid Network) with refinement. 

All NCCT scan slices are resized to $224 \times 224$ to match the pretraining resolution. Data augmentation includes random flipping and rotation. For scans containing skull, windowing in the range of 0--100 HU is applied and rescaled. All datasets are standardized within each task to mitigate acquisition-related biases.

Models are optimized with Adam, using a learning rate of $1 \times 10^{-3}$ for classification and $1 \times 10^{-4}$ for segmentation. Weighted cross-entropy is used for imbalanced classification, while segmentation employs a hybrid loss of binary cross-entropy and dice loss. Training runs for 100 epochs with \textit{ReduceLROnPlateau} scheduling. The batch size is fixed at 8 for both classification and segmentation, and random seed is set to 42 for reproducibility. 

All implementations are based on PyTorch and executed on NVIDIA RTX 4090 GPUs.

\subsection{Results}
\negthinspace

This section reports the performance of our DINOv3-based framework across the main NCCT-based stroke-related tasks. We first summarize quantitative metrics and then provide task-specific insights, focusing on classification and segmentation results. In this section, we don't report many performance results of other models. We only conducted experiments based on DINOv3. We will gradually improve it and give more comparison in our subsequent work. In the following tables, the best matrices will be marked in bold if any comparison include.

\begin{table*}
\caption{Quantitative Results of HeadCT Classification}
\label{tab_1}
\centering
\tabcolsep=0.016\linewidth
\begin{tabular}{ccccccccc}
\hline
Method & AUC & ACC & SEN & SPE  \\
\hline
Anjum et al.\cite{Anjum_HeadCT} & / & \textbf{0.9667} & \textbf{0.9708} & \textbf{0.9625} \\
DINOv3-vitb  & 0.9700 & 0.9300 & 0.9400 & 0.9200 \\
\hline
\end{tabular}
\end{table*}

\subsubsection{Classification Results}
\negthinspace

\begin{figure}[htbp]
    \centering
    \begin{subfigure}[b]{0.48\textwidth}
        \centering
        \includegraphics[width=\textwidth]{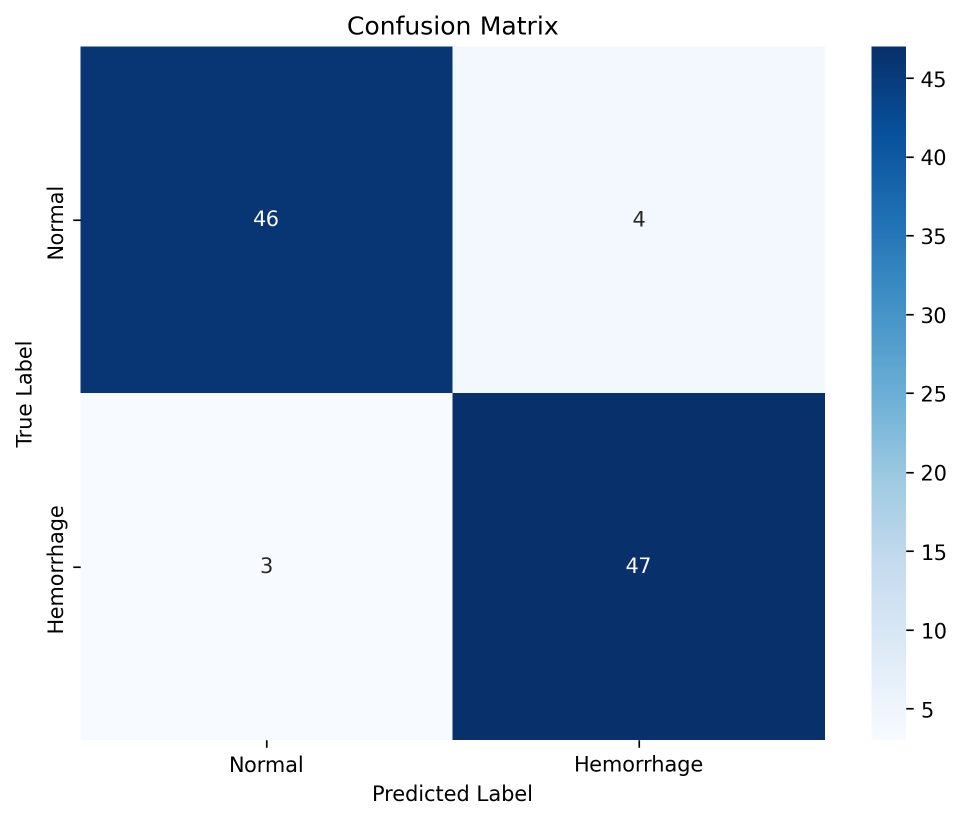}
        \caption{Normal vs. Hemorrhage Confusion Matrix}
        \label{fig:2class}
    \end{subfigure}
    \hfill
    \begin{subfigure}[b]{0.48\textwidth}
        \centering
        \includegraphics[width=\textwidth]{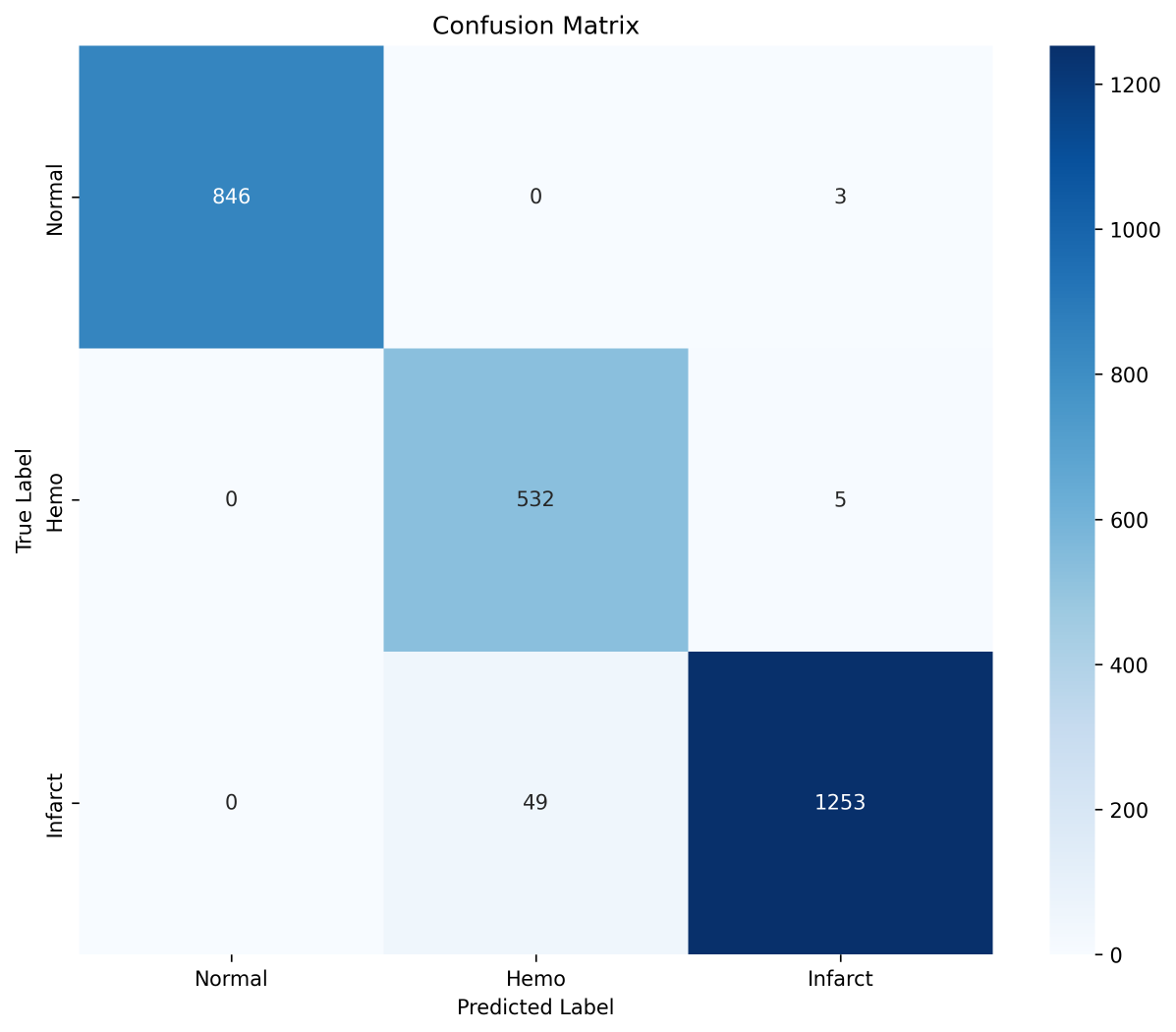}
        \caption{Normal, Hemorrhage, and Infarct Confusion Matrix}
        \label{fig:3class}
    \end{subfigure}
    \caption{Confusion matrices for classification tasks.}
    \label{fig:confusion_matrices}
\end{figure}

For anomaly classification, in binary HeadCT Classification (Normal vs. Hemorrhage), 
Table~\ref{tab_1} shows that DINOv3-vitb achieves an AUC of 0.9700 with 0.9300 accuracy, 0.9400 sensitivity, and 0.9200 specificity. Compared with Anjum et al.\cite{Anjum_HeadCT}, which reports slightly higher accuracy on a larger dataset, our method demonstrates competitive performance with a significantly smaller training set, highlighting the strong transferability of DINOv3 features. Another point worth mentioning is that the training data we used only consisted of 100 slices, and we also had 100 test data. However, for Anjum et al.\cite{Anjum_HeadCT}, the total amount of data used is 2400 slices. The data we used is only a subset of this data.

As for Multi-class Classification (Normal, Infarct, Hemorrhage), as shown in Table~\ref{tab_2}, DINOv3 maintains high per-class performance, achieving near-perfect precision and recall for normal tissue, and strong F1 scores for infarct (0.9778) and hemorrhage (0.9517), with a macro-averaged F1 score of 0.9759. Confusion matrices (Figure~\ref{fig:confusion_matrices}) indicate minimal misclassification, particularly between infarct and hemorrhage.

\begin{table*}
\caption{Quantitative Results of Normal/Infarct/Hemorrhage Classification (Per-class and Overall)}
\label{tab_2}
\centering
\tabcolsep=0.016\linewidth
\begin{tabular}{ccccc}
\hline
Method & Class & Precision & Recall & F1 score  \\
\hline
\multirow{4}{*}{DINOv3-vitb}  
  & Normal   & 1.0000 & 0.9965 & 0.9982 \\
  & Infarct  & 0.9937 & 0.9624 & 0.9778 \\
  & Hemo     & 0.9157 & 0.9907 & 0.9517 \\
  & Overall* & 0.9698 & 0.9832 & 0.9759 \\
\hline
\end{tabular}
\begin{tablenotes}
\footnotesize
\item *Overall: Macro-averaged results across all classes.  
\end{tablenotes}
\end{table*}

\begin{figure}[htbp]
    \centering
    \begin{subfigure}[b]{0.48\textwidth}
        \centering
        \includegraphics[width=\textwidth]{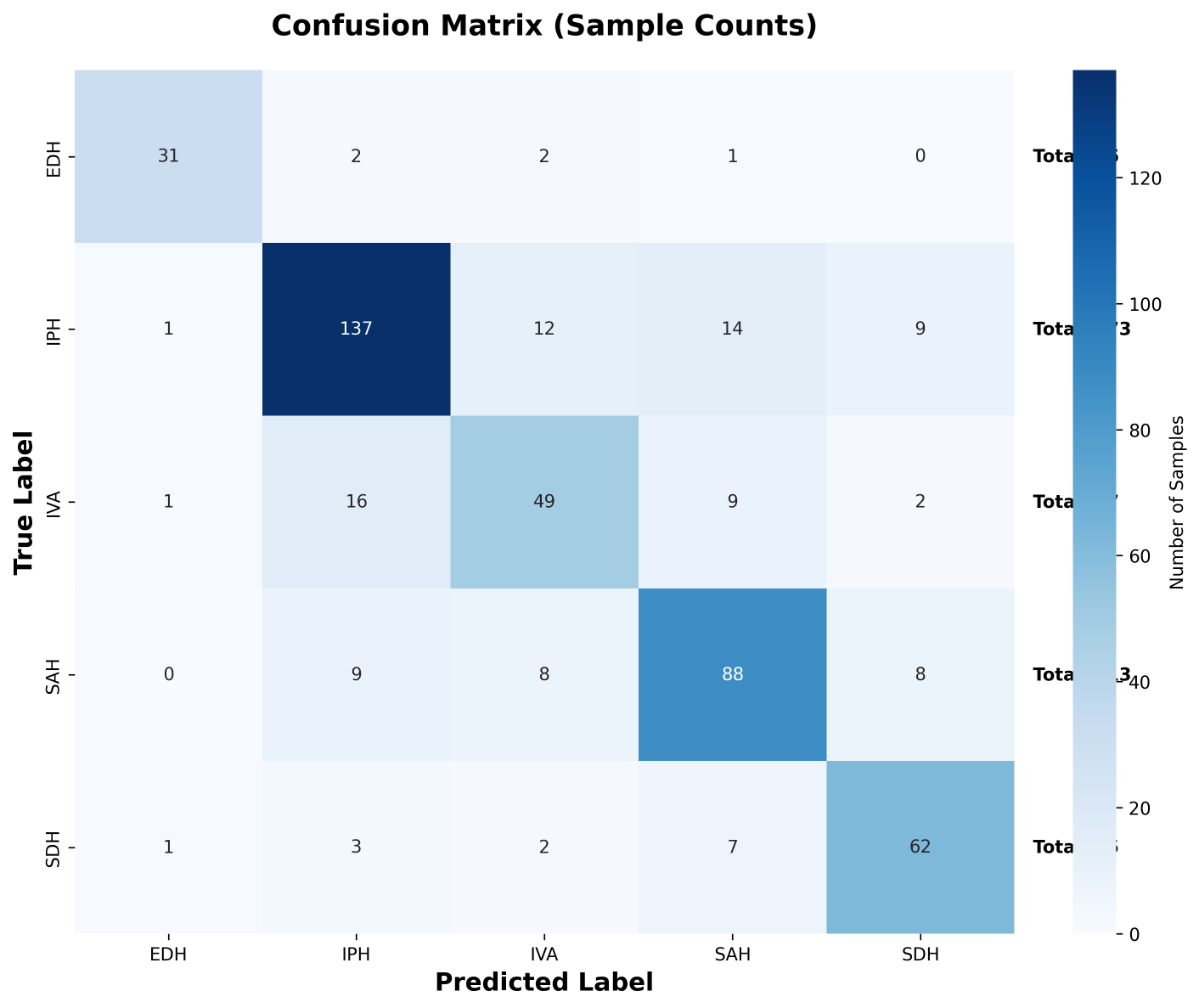}
        \caption{Confusion Matrix Counts}
        \label{fig:bhsd_cls_c}
    \end{subfigure}
    \hfill
    \begin{subfigure}[b]{0.48\textwidth}
        \centering
        \includegraphics[width=\textwidth]{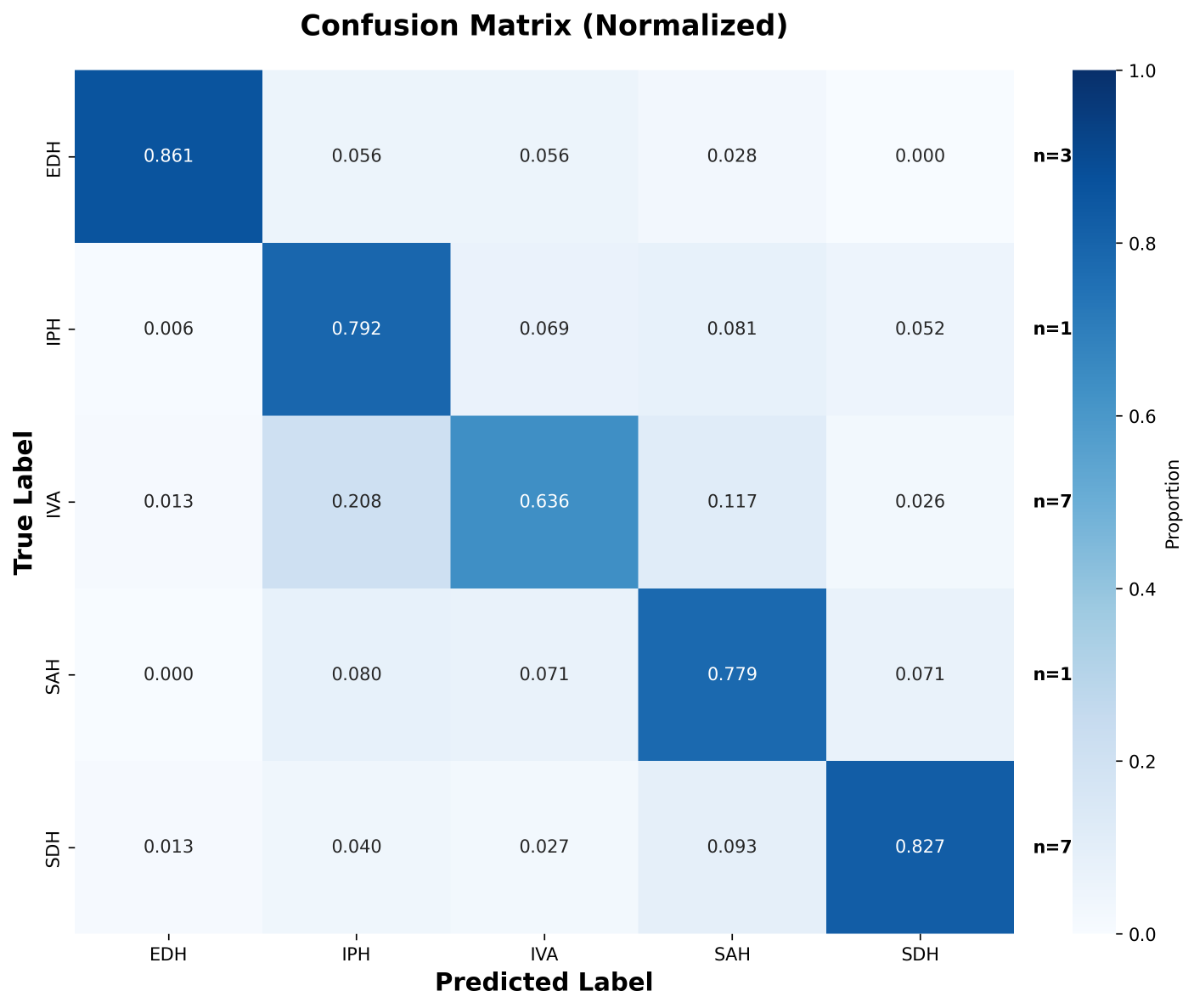}
        \caption{Confusion Matrix Normalized}
        \label{fig:bhsd_cls_n}
    \end{subfigure}
    \caption{Confusion Matrices for Hemorrhage Types Classification}
    \label{fig:bhsd_cls}
\end{figure}

We conduct experiments on hemorrhage subtype classification on BHSD, Table~\ref{tab:bhsd_class_clean} reports per-class and overall performance. DINOv3-vitb achieves strong results for EDH (F1: 0.8857) and SDH (F1: 0.7949), while IVA is more challenging (F1: 0.6533). The overall macro-averaged F1 score of 0.7797 indicates robust discriminative ability across heterogeneous hemorrhage types. Confusion matrices (Figure~\ref{fig:bhsd_cls}) show that most errors occur between clinically similar subtypes. 

\begin{figure}[htbp]
    \centering
    \includegraphics[width=0.96\textwidth]{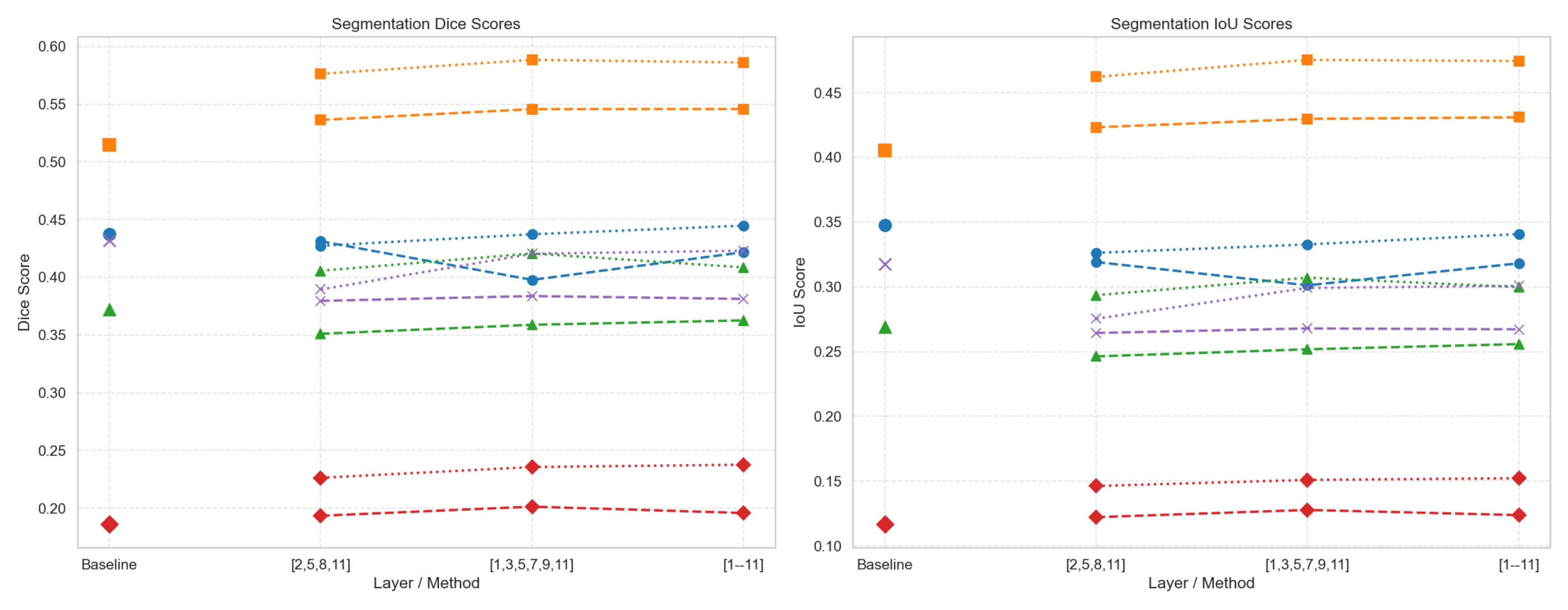}
    \caption{Segmentation performance on the BHSD dataset. Left: Dice scores; Right: IoU scores. Five hemorrhage subtypes are indicated by colors and markers: EDH (blue, circle), IPH (orange, square), IVA (green, triangle), SAH (red, diamond), and SDH (purple, cross). The x-axis represents different methods and layer combinations: Baseline (single-layer decoder), MLP decoder with multiple transformer layers, and FPN+Refine decoder with hierarchical multi-layer feature aggregation. Both Dice and IoU generally improve from Baseline to MLP and further to FPN+Refine, showing the benefit of multi-layer and multi-scale feature integration. Among subtypes, IPH and EDH achieve higher segmentation accuracy, while SAH remains the most challenging due to its heterogeneity.}
    \label{fig:bhsd_seg}
\end{figure}

\begin{table*}[ht]
\caption{Classification results on BHSD dataset using linear probe.}
\label{tab:bhsd_class_clean}
\centering
\begin{tabular}{ccccc}
\hline
Method & Class & Precision & Recall & F1 score  \\
\hline
\multirow{5}{*}{DINOv3-vitb}  
  & EDH & 0.9118 & 0.8611 & 0.8857 \\
  & IPH & 0.8204 & 0.7919 & 0.8059 \\
  & IVA & 0.6712 & 0.6364 & 0.6533 \\
  & SAH & 0.7395 & 0.7788 & 0.7586 \\
  & SDH & 0.7654 & 0.8267 & 0.7949 \\
  & Overall* & 0.7817 & 0.7790 & 0.7797 \\
\hline
\end{tabular}
\begin{tablenotes}
\footnotesize
\item *Overall: Macro-averaged results across all classes.  
\end{tablenotes}
\end{table*}

For our extend experiment for Volumetric ASPECTS Classification, using mean-pooled slice-wise features, the model achieves an AUC of 0.6760 and accuracy of 0.8571 (Table~\ref{tab:private_tasks_full}), suggesting that DINOv3 features capture clinically relevant volumetric patterns even without native 3D modeling. However, due to the limited testing data, we need more experiments and origanized data for validating this task.

\begin{table*}[ht]
\caption{Infarct segmentation results on three datasets: AISD, ISLES24, and private infarct segmentation dataset}
\label{tab:all_seg}
\centering
\begin{tabular}{cccccc}
\hline
Dataset & Method & Layers & IoU & Dice & HD95 \\
\hline
\multirow{7}{*}{AISD} 
 & Baseline & [11] & 0.2067 & 0.2916 & 51.6977 \\
 & MLP & [2,5,8,11] & 0.2457 & 0.3332 & \textbf{40.9900} \\
 & MLP & [1,3,5,7,9,11] & 0.2459 & 0.3371 & 43.6547 \\
 & MLP & [1--11] & 0.2488 & 0.3377 & 43.0443 \\
 & FPN+Refine & [2,5,8,11] & 0.2629 & 0.3637 & 48.6337 \\
 & FPN+Refine & [1,3,5,7,9,11] & \textbf{0.2703} & \textbf{0.3674} & 43.8991 \\
 & FPN+Refine & [1--11] & 0.2508 & 0.3415 & 46.8756 \\
\hline
\multirow{7}{*}{ISLES24} 
 & Baseline & [11] & 0.2848 & 0.3849 & 37.1890 \\
 & MLP & [2,5,8,11] & 0.4090 & 0.5216 & 23.5562 \\
 & MLP & [1,3,5,7,9,11] & 0.4098 & 0.5214 & 23.9581 \\
 & MLP & [1--11] & 0.4173 & 0.5314 & 23.7450 \\
 & FPN+Refine & [2,5,8,11] & \textbf{0.4507} & \textbf{0.5678} & 22.2160 \\
 & FPN+Refine & [1,3,5,7,9,11] & 0.4428 & 0.5583 & \textbf{21.9963} \\
 & FPN+Refine & [1--11] & 0.4363 & 0.5528 & 23.8718 \\
\hline
\multirow{7}{*}{Private infarct} 
 & Baseline & [11] & 0.4416 & 0.5654 & 36.2112 \\
 & MLP & [2,5,8,11] & 0.4805 & 0.6024 & 30.2230 \\
 & MLP & [1,3,5,7,9,11] & 0.4796 & 0.6004 & 31.9453 \\
 & MLP & [1--11] & 0.4868 & 0.6072 & \textbf{29.0593} \\
 & FPN+Refine & [2,5,8,11] & 0.4986 & 0.6177 & 30.6560 \\
 & FPN+Refine & [1,3,5,7,9,11] & \textbf{0.5169} & \textbf{0.6371} & 29.9949 \\
 & FPN+Refine & [1--11] & 0.4972 & 0.6197 & 32.2705 \\
\hline
\end{tabular}
\end{table*}

\begin{table*}[ht]
\caption{Segmentation results on IHDS and Private Hemorrhage datasets. Best Dice/IoU (higher better) and HD95 (lower better) are highlighted in bold.}
\label{tab:ihds_hemo}
\centering
\begin{tabular}{lccc|ccc}
\hline
& \multicolumn{3}{c|}{IHDS} & \multicolumn{3}{c}{Private} \\
Method & Dice & IoU & HD95 & Dice & IoU & HD95 \\
\hline
Upsampling & 0.6135 & 0.4669 & 12.6802 & 0.7893 & 0.6860 & 12.4490 \\
MLP [2,5,8,11] & 0.6251 & 0.4807 & 12.5267 & 0.7907 & 0.6913 & 12.4102 \\
FPN+Refine [2,5,8,11] & \textbf{0.6363} & \textbf{0.4959} & 12.2503 & \textbf{0.7936} & \textbf{0.6942} & \textbf{12.2643} \\
FPN+Refine [1,3,5,7,9,11] & 0.6352 & 0.4956 & \textbf{12.2503} & 0.7912 & 0.6921 & 12.2894 \\
\hline
\end{tabular}
\end{table*}

\subsubsection{Segmentation Results}
\negthinspace

For infarct segmentation, results on AISD, ISLES24, and a private dataset (Table~\ref{tab:all_seg}) show that multi-layer feature integration consistently improves performance compared with single-layer baselines. On AISD, the best Dice of 0.3674, IoU of 0.2703, and HD95 of 40.9900 are achieved by FPN+Refine with layers [1,3,5,7,9,11]. On ISLES24, FPN+Refine with layers [2,5,8,11] yields the highest Dice of 0.5678, IoU of 0.4507, and the lowest HD95 of 21.9963. On the private dataset, FPN+Refine with layers [1,3,5,7,9,11] achieves the best Dice of 0.6371, IoU of 0.5169, and HD95 of 29.0593.

\begin{table*}[ht]
\centering
\caption{Results on private dataset for volumetric tasks, including train/test split and class distribution using mean arregation.}
\label{tab:private_tasks_full}
\begin{tabular}{>{\centering\arraybackslash}m{2.5cm} l >{\centering\arraybackslash}m{2.5cm} l} 
\hline
Task & Result & Train / Test & Class Distribution \\
\hline
Segmentation & 
\begin{tabular}[c]{@{}l@{}}
Dice: 0.2669 $\pm$ 0.1840 \\
IoU: 0.1674 $\pm$ 0.1300
\end{tabular} & 200 / 49 & - \\
\hline
Classification (Aspects) & 
\begin{tabular}[c]{@{}l@{}}
AUC: 0.6760 \\
Accuracy: 0.8571 \\
Sensitivity: 0.8929 \\
Specificity: 0.5714
\end{tabular} & 250 / 63 & 
\begin{tabular}[c]{@{}l@{}}
Train: Label 0: 27\\Label 1: 223\\
Test: Label 0: 7\\Label 1: 56
\end{tabular} \\
\hline
\end{tabular} \\
\end{table*}

And for hemorrhage Segmentation, we evaluate models on IHDS, BHSD, and a private dataset (Table~\ref{tab:ihds_hemo}, Figure~\ref{fig:bhsd_seg}). 
On IHDS, FPN+Refine with layers [2,5,8,11] yields the best Dice (0.6363) and IoU (0.4959), while the lowest HD95 (12.2503) is achieved by FPN+Refine with layers [1,3,5,7,9,11]. 
On the private dataset, FPN+Refine again achieves the highest Dice (0.7936) and IoU (0.6942), with the lowest HD95 (12.2643) from [2,5,8,11]. 
On BHSD, subtype-level analysis (Figure~\ref{fig:bhsd_seg}) shows that multi-layer integration via FPN+Refine consistently improves performance compared with both the baseline and MLP-only decoders. 
Among subtypes, intraparenchymal hemorrhage (IPH) achieves the highest segmentation accuracy, with Dice improving from 0.5148 (baseline) to 0.5884 (FPN+Refine). 
Epidural hemorrhage (EDH) also benefits notably (Dice up to 0.4448), while subarachnoid hemorrhage (SAH) remains the most challenging, reaching only 0.2377 Dice despite improvement from the baseline (0.1861). 
These results highlight that while multi-scale feature integration is broadly beneficial, the segmentation difficulty strongly depends on hemorrhage subtype heterogeneity and anatomical presentation.

In our private volumetric lesion segmentation dataset, Aggregating slice-wise features yields a Dice score of 0.2669 $\pm$ 0.1840 and IoU of 0.1674 $\pm$ 0.1300 (Table~\ref{tab:private_tasks_full}). Although performance is limited by the lack of direct spatial contextual information, results highlight the potential of leveraging DINOv3 features for volumetric inference.

\section{Discussion and Conclusion}
\negthinspace

This work presents a unified, reproducible evaluation of a frozen self-supervised transformer backbone (DINOv3) on NCCT-based stroke analysis spanning detection, subtyping, and lesion segmentation in different settings. Our study yields two main insights.

First, high-fidelity dense features learned from large-scale self-supervision transfer effectively to low-contrast NCCT. With minimal task-specific parameters and limited labeled data, a frozen DINOv3-ViT-B achieves strong performance on binary hemorrhage detection and three-class (normal/infarct/hemorrhage) classification, and provides competitive signals for fine-grained hemorrhage subtyping on BHSD. These results suggest that self-supervised transformers can serve as robust backbones for NCCT even under label scarcity and device heterogeneity.

Second, multi-layer feature integration is crucial for pixel-level prediction on subtle NCCT findings. For infarct and hemorrhage segmentation, fusing intermediate transformer layers with a DPT-style (FPN+Refine) decoder consistently improves Dice/IoU over single-layer baselines, highlighting that hierarchical transformer features jointly encode global context and local structure needed for delineating low-contrast lesions.

Limitations and future work: Our frozen-backbone protocol isolates representation quality but may underutilize domain-specific cues; mild fine-tuning, domain-adaptive normalization, or adapter-based updates could strengthen volumetric tasks. For 3D segmentation, hybrid 2.5D contexts, cross-slice attention, or lightweight 3D decoders with learned skip surrogates may close the gap. Extending benchmarks to additional public NCCT datasets, reporting confidence intervals and statistical tests, and releasing pretrained heads/splits will further enhance reproducibility and clinical relevance. Finally, integrating clinical variables (e.g., NIHSS, time-to-scan) and multimodal data (CTA/CTP) could advance prognostic modeling beyond NCCT alone.

In summary, we establish a practical baseline for applying self-supervised transformer features to NCCT-based stroke analysis. Our results demonstrate that DINOv3’s high-fidelity dense representations support strong performance in 2D classification and segmentation, enable systematic study of multi-layer fusion on low-contrast lesions, and provide actionable baselines for slice-aggregated 3D tasks. We release unified protocols, metrics, and dataset to facilitate fair comparison and to accelerate future developments in NCCT-centered stroke AI.

%
%
%
 \bibliographystyle{splncs04}
 \bibliography{mybib.bib}

\begin{thebibliography}{10}
\providecommand{\url}[1]{\texttt{#1}}
\providecommand{\urlprefix}{URL }
\providecommand{\doi}[1]{https://doi.org/#1}

\bibitem{Anjum_HeadCT}
Anjum, N., Sakib, A.N.M., Masudul~Ahsan, S.M.: Classification of brain hemorrhage using deep learning from ct scan images. In: Ahmad, M., Uddin, M.S., Jang, Y.M. (eds.) Proceedings of International Conference on Information and Communication Technology for Development. pp. 181--193. Springer Nature Singapore, Singapore (2023)

\bibitem{visionR10834497}
Awais, M., Naseer, M., Khan, S., Anwer, R.M., Cholakkal, H., Shah, M., Yang, M.H., Khan, F.S.: Foundation models defining a new era in vision: A survey and outlook. IEEE Transactions on Pattern Analysis and Machine Intelligence  \textbf{47}(4),  2245--2264 (2025). \doi{10.1109/TPAMI.2024.3506283}

\bibitem{visionR3}
Azad, B., Azad, R., Eskandari, S., Bozorgpour, A., Kazerouni, A., Rekik, I., Merhof, D.: Foundational models in medical imaging: A comprehensive survey and future vision (2023), \url{https://arxiv.org/abs/2310.18689}

\bibitem{pmlr-v119-chen20j}
Chen, T., Kornblith, S., Norouzi, M., Hinton, G.: A simple framework for contrastive learning of visual representations. In: III, H.D., Singh, A. (eds.) Proceedings of the 37th International Conference on Machine Learning. Proceedings of Machine Learning Research, vol.~119, pp. 1597--1607. PMLR (13--18 Jul 2020), \url{https://proceedings.mlr.press/v119/chen20j.html}

\bibitem{Eswaradass02082016}
Eswaradass, P., Appireddy, R., Evans, J., Tham, C., Dey, S., Najm, M., Menon, B.K.: Imaging in acute stroke. Expert Review of Cardiovascular Therapy  \textbf{14}(8),  963--975 (2016). \doi{10.1080/14779072.2016.1196134}, pMID: 27249030

\bibitem{he2021masked}
He, K., Chen, X., Xie, S., Li, Y., Dollár, P., Girshick, R.: Masked autoencoders are scalable vision learners (2021)

\bibitem{Hssayeni2020CTHemorrhage}
Hssayeni, M.: Computed tomography images for intracranial hemorrhage detection and segmentation (version 1.3.1) (2020). \doi{10.13026/4nae-zg36}, \url{https://doi.org/10.13026/4nae-zg36}, rRID:SCR\_007345

\bibitem{data5010014}
Hssayeni, M.D., Croock, M.S., Salman, A.D., Al-khafaji, H.F., Yahya, Z.A., Ghoraani, B.: Intracranial hemorrhage segmentation using a deep convolutional model. Data  \textbf{5}(1) (2020). \doi{10.3390/data5010014}, \url{https://www.mdpi.com/2306-5729/5/1/14}

\bibitem{liang2021SymmetryEnhancedAN}
Liang, K., Han, K., Li, X., Cheng, X., Li, Y., Wang, Y., Yu, Y.: Symmetry-enhanced attention network for acute ischemic infarct segmentation with non-contrast ct images. In: MICCAI (2021)

\bibitem{VisionR10916803}
Lu, S., Guo, J., Zimmer-Dauphinee, J.R., Nieusma, J.M., Wang, X., VanValkenburgh, P., Wernke, S.A., Huo, Y.: Vision foundation models in remote sensing: A survey. IEEE Geoscience and Remote Sensing Magazine  \textbf{13}(3),  190--215 (2025). \doi{10.1109/MGRS.2025.3541952}

\bibitem{NEETHI2022103720}
Neethi, A., Niyas, S., Kannath, S.K., Mathew, J., Anzar, A.M., Rajan, J.: Stroke classification from computed tomography scans using 3d convolutional neural network. Biomedical Signal Processing and Control  \textbf{76},  103720 (2022). \doi{https://doi.org/10.1016/j.bspc.2022.103720}, \url{https://www.sciencedirect.com/science/article/pii/S1746809422002427}

\bibitem{strokeReview1}
Patil, S., Rossi, R., Jabrah, D., Doyle, K.: Detection, diagnosis and treatment of acute ischemic stroke: Current and future perspectives. Frontiers in Medical Technology  \textbf{Volume 4 - 2022} (2022). \doi{10.3389/fmedt.2022.748949}

\bibitem{radford2021learning}
Radford, A., Kim, J.W., Hallacy, C., Ramesh, A., Goh, G., Agarwal, S., Sastry, G., Askell, A., Mishkin, P., Clark, J., Krueger, G., Sutskever, I.: Learning transferable visual models from natural language supervision (2021)

\bibitem{Riedel2024ISLES}
Riedel, O.E., de~la Rosa, E., Hernandez~Petzsche, M., Baazaoui, H., Yang, K., Musio, F.A., Kirschke, J.S.: Isles'24 -- a real-world longitudinal multimodal stroke dataset (2024), \url{https://arxiv.org/abs/2408.11142}

\bibitem{unet}
Ronneberger, O., Fischer, P., Brox, T.: U-net: Convolutional networks for biomedical image segmentation. arXiv:1505.04597 [cs.CV] (2015)

\bibitem{Rosa2024ISLES}
de~la Rosa, E., Su, R., Reyes, M., Wiest, R., Riedel, E.O., Kofler, F., Menze, B.: Isles'24: Final infarct prediction with multimodal imaging and clinical data. where do we stand? (2024), \url{https://arxiv.org/abs/2408.10966}

\bibitem{simeoni2025dinov3}
Sim{\'e}oni, O., Vo, H.V., Seitzer, M., Baldassarre, F., Oquab, M., Jose, C., Khalidov, V., Szafraniec, M., Yi, S., Ramamonjisoa, M., Massa, F., Haziza, D., Wehrstedt, L., Wang, J., Darcet, T., Moutakanni, T., Sentana, L., Roberts, C., Vedaldi, A., Tolan, J., Brandt, J., Couprie, C., Mairal, J., J{\'e}gou, H., Labatut, P., Bojanowski, P.: {DINOv3} (2025), \url{https://arxiv.org/abs/2508.10104}

\bibitem{11078788}
Soomro, T.A., Ali, A., Faye, I., Memon, K.A., Alothman, A., Mahmud, R.A., Muda, A.S.B., Qureshi, K.K.: Improving stroke diagnosis and treatment with deep learning: A review. IEEE Sensors Journal  \textbf{25}(16),  30275--30290 (2025). \doi{10.1109/JSEN.2025.3586157}

\bibitem{su2021roformer}
Su, J., Lu, Y., Pan, S., Murtadha, A., Wen, B., Liu, Y.: Roformer: Enhanced transformer with rotary position embedding (2021)

\bibitem{UMAMAHESWARAN2024122559}
UmaMaheswaran, S., Ahmad, F., Hegde, R., Alwakeel, A.M., {Rameem Zahra}, S.: Enhanced non-contrast computed tomography images for early acute stroke detection using machine learning approach. Expert Systems with Applications  \textbf{240},  122559 (2024). \doi{https://doi.org/10.1016/j.eswa.2023.122559}, \url{https://www.sciencedirect.com/science/article/pii/S0957417423030610}

\bibitem{WANG2025108488}
Wang, X., Meng, Y., Dong, Z., Cao, Z., He, Y., Sun, T., Zhou, Q., Niu, G., Ding, Z., Shi, F., Shen, D.: Segmentation of infarct lesions and prognosis prediction for acute ischemic stroke using non-contrast ct scans. Computer Methods and Programs in Biomedicine  \textbf{258},  108488 (2025). \doi{https://doi.org/10.1016/j.cmpb.2024.108488}, \url{https://www.sciencedirect.com/science/article/pii/S0169260724004814}

\bibitem{wu2023bhsd}
Wu, B., Xie, Y., Zhang, Z., Ge, J., Yaxley, K., Bahadir, S., Wu, Q., Liu, Y., To, M.S.: Bhsd: A 3d multi-class brain hemorrhage segmentation dataset. In: International Workshop on Machine Learning in Medical Imaging. pp. 147--156. Springer (2023)

\bibitem{yang2025segdino}
Yang, S., Wang, H., Xing, Z., Chen, S., Zhu, L.: Segdino: An efficient design for medical and natural image segmentation with dino-v3. arXiv preprint arXiv:2509.00833  (2025)

\bibitem{ZHANG2023110959}
Zhang, L., Wu, J., Yu, R., Xu, R., Yang, J., Fan, Q., Wang, D., Zhang, W.: Non-contrast ct radiomics and machine learning for outcomes prediction of patients with acute ischemic stroke receiving conventional treatment. European Journal of Radiology  \textbf{165},  110959 (2023). \doi{https://doi.org/10.1016/j.ejrad.2023.110959}, \url{https://www.sciencedirect.com/science/article/pii/S0720048X23002735}

\bibitem{zhou2022ibot}
Zhou, J., Wei, C., Wang, H., Shen, W., Xie, C., Yuille, A., Kong, T.: ibot: Image bert pre-training with online tokenizer. In: International Conference on Learning Representations (ICLR) (2022), \url{https://arxiv.org/abs/2111.07832}

\end{thebibliography}

\end{document}